\documentclass[9pt,journal]{IEEEtran}
\twocolumn
\usepackage{graphicx}
\usepackage{caption}
\usepackage{subcaption}
\usepackage{amsfonts}
\usepackage[centertags]{amsmath}
\usepackage{amssymb}
\usepackage{floatrow}
\newfloatcommand{capbtabbox}{table}[][\FBwidth]

\usepackage{blindtext}
\newtheorem{theorem}{Theorem}[section]
\newtheorem{lemma}[theorem]{Lemma}

\usepackage{bm}
\usepackage{amsmath}
\usepackage{amsfonts}
\usepackage{amssymb}
\usepackage{amscd}

\usepackage{newlfont}
\newcommand{\xbf}{\mathbf{x}}

\newcommand{\ybf}{\mathbf{y}}

\usepackage{float}
\usepackage{colortbl}
\ifCLASSINFOpdf
\else
\fi

\begin{document}
\title{Type I and Type II Bayesian Methods for Sparse Signal Recovery  using Scale Mixtures}

\author{Ritwik Giri and
        Bhaskar Rao
  %      and~Jane~Doe,~\IEEEmembership{Life~Fellow,~IEEE}% <-this % stops a space
\thanks{Authors are members of DSP lab (dsp.ucsd.edu), Electrical and Computer Engineering, University of California, San Diego,
CA, 92122 USA
 e-mail: rgiri@ucsd.edu, brao@ucsd.edu}}% <-this % stops a space
\maketitle

\begin{abstract}
In this paper, we propose a generalized scale mixture family of distributions, namely the Power Exponential Scale Mixture (PESM) family, 
to model the sparsity inducing priors currently in use for sparse signal recovery (SSR).  We show that the successful 
and popular methods such as LASSO, Reweighted  $\ell_1$ and Reweighted $\ell_2$ methods can be formulated in an 
unified manner in a maximum a posteriori (MAP) or Type I Bayesian framework using an appropriate member of the 
PESM family as the sparsity inducing prior. In addition, exploiting the natural hierarchical framework induced 
by the PESM family, we utilize these priors in a Type II framework and develop the corresponding EM based estimation 
algorithms.  Some insight into the differences between Type I and Type II methods is provided and of particular 
interest in the algorithmic development is the Type II variant of the popular and successful reweighted $\ell_1$ method. 
Extensive empirical results are provided and they show that the Type II methods exhibit better support recovery than the corresponding Type I methods.

\end{abstract}

\begin{IEEEkeywords}
Sparse Bayesian Learning, LASSO, Reweighted $\ell_1$, Reweighted $\ell_2$, Gaussian Scale Mixture
\end{IEEEkeywords}

% \fi
\IEEEpeerreviewmaketitle

\section{Introduction}

Sparse signal recovery (SSR), i.e, finding sparse signal representations from overcomplete dictionaries, has become a very active research area in recent times because of its wide range engineering applications and interesting theoretical nature \cite{eldar2012compressed, bruckstein2009sparse, donoho2006compressed, elad2010sparse}.

\subsection{Problem Formulation}

The SSR problem involves solving an under-determined system of equations $\mathbf{y}= \Phi \mathbf{x},$
where vector $\mathbf{y}$ is the
$N \times 1$ measurement vector and $\Phi$ is the $N \times M $ overcomplete dictionary, where $M > N, $and it is
assumed that $\mbox{Spark}({\mathbf{\Phi}}) = N.$ $\mathbf{\Phi}$ are often formed from a physically meaningful model and the
vector $\xbf$ is the
$M \times 1$  vector of interest.
As the system has fewer equations than unknowns, it can have infinitely many
solutions and thus additional information is needed to identify which of these candidate
solutions is indeed the appropriate one for the problem at hand. In the SSR problem, it will be assumed that the solution of interest is sparse, i.e. most of the entries are zero.
Ideally one can recover the optimal sparsest solution $\mathbf{x}_0$ by solving the following $\ell_0$ optimization problem \cite{elad2010sparse},
\begin{equation}
\min_{x} ||\mathbf{x}||_0 \; \; \text{such that} \;\mathbf{y}= \Phi \mathbf{x},
\end{equation}
where $\|\xbf\|_0$ is a measure of the support of $\xbf.$
In practice, measurements are generally corrupted by noise, which motivates  the following modified optimization problem:
\begin{equation}
\min_x ||\mathbf{y} - \Phi \mathbf{x}||_2^2 + \lambda ||\mathbf{x}||_0
\end{equation}
where, $\lambda > 0$ is related to the  measurement noise variance.
%It can be shown that in the limit as $\lambda \to 0$, the above two problems are equivalent.

However, the above optimization problem is not convex and is known to be NP-hard. For computational tractability, the original penalty factor $||\mathbf{x}||_0$ is often approximated by a suitable surrogate $g(\mathbf{x})$ leading to the optimization problem
\begin{equation}
\min_x ||\mathbf{y} - \Phi \mathbf{x}||_2^2 + \lambda g(\mathbf{x})
\label{l1}
\end{equation}
Different choices of the penalty factor $g(\mathbf{x}),$ also referred to here as diversity measure, lead to different SSR algorithms \cite{candes2008enhancing, chartrand2008iteratively, daubechies2010iteratively, donoho2003optimally}, and it has been shown that the choice of a strictly concave penalty factor on the positive orthant leads to a objective function with local minima being sparse and sparsest solution as a global minimum under some conditions \cite{rao2003subset}. Majorization-Minimization \cite{figueiredo2007majorization} can be employed to solve this optimization problem for such penalty functions, and this has led to the development of useful and effective reweighted norm minimization algorithms. Typically, $\ell_1$ and $\ell_2$ norms are selected because of their convex nature and the later because of the closed form solution at each iteration.

\subsection{Related Literature}
Minimizing diversity measures $g(\xbf)$ to recover the sparse representations has been a popular algorithm exploration avenue. In this framework, the SSR problem formulation can also be viewed as a  regularization approach to signal
 reconstruction. A popular approach among this class is the $\ell_p$ norm minimization based methods. $p=1$ leads to a tractable and computationally attractive convex optimization problem and  the very well known approaches such as Basis Pursuit, LASSO are based on the $\ell_1$ framework \cite{tibshirani1996regression, donoho2006most}. Other than the convexity property, $\ell_1$ based approaches have been supported by theoretical guarantees of exact recovery given some conditions on the overcomplete dictionary \cite{donoho2003optimally}, which makes these approaches attractive options. The recently proposed reweighted $\ell_1$ and $\ell_2$ norm minimization approaches \cite{chartrand2008iteratively, candes2008enhancing,rao1999affine}  have empirically shown superior recovery performance over $\ell_1$ minimization and are considered in this work.

In addition to the regularization framework, another options for SSR algorithm development is the Bayesian framework \cite{he2009exploiting, ji2008bayesian,babacan2010bayesian, vila2014empirical, vila2013expectation, baron2010bayesian, asilomar}. In a Bayesian framework, the sparsity constraint is incorporated by choosing a suitable sparse prior on the coefficient vector $\mathbf{x}$. In a Bayesian setting, there are two popular
avenues for algorithm development: a Type I MAP based approach, and a Type II Evidence Maximization approach involving a Hierarchical model. Most of the approaches discussed above, based on  (\ref{l1}), can be interpreted and cast in a suitable Type I framework. A Type II framework has been considered in \cite{tipping2001sparse, ji2008bayesian}, where a Relevance Vector Machine is adapted to the problem at hand. In \cite{wipf2010iterative, wipf2008new, wu2012dual} a Type II optimization problem has been transformed into a Type I problem by employing a suitable penalty function and reweighted norm minimization algorithm is developed to solve the
resulting optimization problem. Following the Type II framework, a Laplacian prior which corresponds to $\ell_1$ norm minimization can also be represented in a Hierarchy using a Gaussian Scale Mixture (GSM) representation \cite{babacan2010bayesian, figueiredo2003adaptive}. In the statistics community, the well known Bayesian Lasso \cite{park2008bayesian}  also makes use of the equivalence of a hierarchical Gaussian-Exponential
prior to the Laplace prior, and conducts a fully Bayesian inference (via Markov
chain Monte Carlo or MCMC sampling algorithms). Demi-Bayesian Lasso \cite{balakrishnan2009priors} solves the same problem using a Type II approach. It has been shown empirically that a Type II methods performs consistently better than Type I, i.e the MAP estimation approach, and theoretical analysis in support for this superiority has recently begun to appear. However, much remains to be done and this work is an attempt in this direction. In \cite{mackay1999comparison}, the two different frameworks are analyzed in a generalized Hierarchical Bayesian setting which motivates us to analyze these two frameworks for the specific SSR problem to gather additional insights by exploiting domain knowledge. In \cite{palmer2005variational}, Type I and Type II frameworks for SSR were introduced using two forms of density representation, a convex representation and a GSM representation, to provide an unified treatment. We build on this work and 
employ a generalized scale mixture representation to establish connections and develop enhancements to popular SSR algorithms, as well as treat both $\ell_1$ and $\ell_2$ variants in an unified manner.

As mentioned above, a key ingredient behind the Type II methods is the Scale Mixture/Hierarchical representation of the super gaussian priors, which allows one to design efficient algorithms conveniently. Gaussian Scale Mixtures (GSM) \cite{liu1995ml,lange1993normal, palmer2005variational}  and Laplacian Scale mixtures (LSM)  \cite{garrigues2010group} have been studied before  in the context of sparsity. In this work we discuss a more general Scale Mixture framework, the Power Exponential Scale Mixture (PESM) family, for SSR algorithm development. The PESM representation includes the popular GSM and LSM as special cases and provides a mechanism to provide a unified view of the popular $\ell_1$ and $\ell_2$ frameworks currently employed. This work will emphasize the generalized t (GT) distribution family of priors, a member of PESM, since it has a wide range of tail shapes, and also includes the heavy tailed super gaussian distributions. GT family of distributions have been mentioned in statistics literatures for design of robust regressors for several financial modeling tasks, where the heavy tail nature of GT helps to model the outliers \cite{mcdonald1988partially, butler1990robust}.

 \subsection{Contributions of the Paper}

 \begin{itemize}
\item
We discuss a generalized scale mixture framework, the power exponential scale mixture (PESM) family, and show how many of the super gaussian densities used in practice can be represented using this framework.
 \item
 We summarize two types of Bayesian frameworks, i.e. Type I and Type II for SSR in detail, along with providing connections to traditional norm minimization approaches by suitable choice of sparse prior distributions.
 Of particular importance is the treatment of the diversity measure used in connection with the reweighted $\ell_1$ algorithm as well as an unified treatment of both $\ell_1$ and $\ell_2$ based approaches.

\item
We formulate and unify three well known diversity minimization based SSR algorithms in the PESM  framework and derive the Type I and Type II versions of them. Of particular interest is the Type II counterpart of the
reweighted $\ell_1$ algorithm \cite{candes2008enhancing}.

\item
 We analyze the difference between Type I and Type II inference procedures and our analysis shows the fundamental difference between these two frameworks and also helps to understand a potential reason for the empirical superiority of Type II methods over Type I.

\item
 Extensive empirical experimentation results are presented to support the superiority claim of Type II methods over their Type I counterpart.
 \end{itemize}

\subsection{Article Organization}

The rest of the paper is organized as follows. In Section~\ref{sec:PESM}, a generalized scale mixture representation, the Power Exponential Scale Mixtures (PESM) family, is presented which are of main importance to design Bayesian methods for SSR. In Section III we discuss Type I/MAP algorithms for SSR and derive a unified inference procedure and provide connection with three well known SSR algorithms.  In Section IV we discuss Type II framework for SSR along with analyzing the fundamental difference between Type I and Type II algorithms. The EM based inference procedure for Type II algorithms to estimate the coefficient vector and the hyperparameters is also developed which includes
the counterpart to the popular reweighted $\ell_1$ method. We present experimental results of the proposed algorithms in Section V in different settings and finally conclusions are presented in Section VI.

\section{Scale Mixture Distributions}

Scale mixture distributions namely Gaussian Scale mixtures (GSM) and Laplacian Scale mixtures (LSM) have gained lot of attention in recent years because of their ability to represent complex heavy tailed super gaussian distributions in a simple hierarchical manner \cite{liu1995ml,lange1993normal, palmer2005variational, garrigues2010group}. In the statistics community,  robustness has been the major reason for the use of scale mixtures. In regression analysis, the method of least squares often fails because of the outliers in the data. The need to model the outliers motivates the use of heavy tailed distribution.

\subsection{Power Exponential Scale Mixture (PESM) distribution}
\label{sec:PESM}
In this work,
 a more general Power Exponential Scale Mixture (PESM) distribution, which is a generalization of GSM and LSM, is presented.
The PESM representation is then used to model the prior sparse distribution over the  vector $\xbf$ and for sparse signal recovery algorithm development.

Power exponential (PE) distributions were first introduced by Box and Tiao (1962) in the context of robust regression to deal with non-normality. PE distribution is symmetric about the origin and a zero mean PE distribution has the following parameterized form:
\begin{equation}
PE(x; p, \sigma) = \frac{p \;e^{(-\frac{|x|}{\sigma})^p}}{2 \sigma \Gamma(\frac{1}{p})}
\end{equation}

It is evident from the above given form, that $p=2$ results in the normal distribution, whereas $p=1$ connects to the well known Double exponential or Laplacian distribution. $p <2$ leads to distribution with heavier tails than the Gaussian distribution.

\begin{figure}
\centering
  \includegraphics[width=\linewidth,height=1.8in]{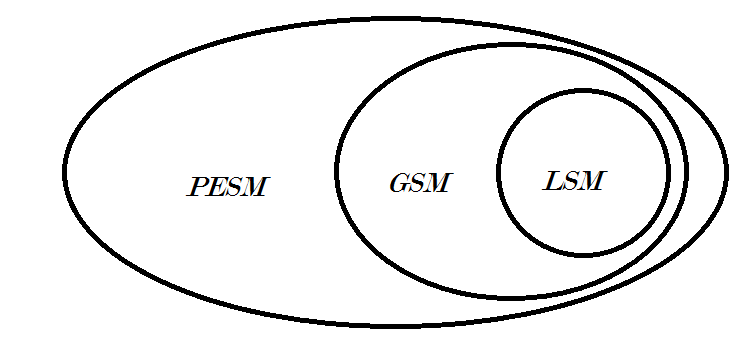}
  \caption{PESM: Generalized scale mixtures}
\label{fig:venn}
\end{figure}

PESM family of distributions refer to distributions that can be represented as follows:
%Sparse prior over the coefficient vector $x$ can be represented in a hierarchy using scale mixture of PE distribution.
\begin{equation}
p(x)= \int p( x| \gamma) p(\gamma) d\gamma = \int PE( x;   p, \gamma) p(\gamma) d\gamma
\label{eq:mix}
\end{equation}
%Now, in a bayesian framework sparsity on coefficient vector is imposed by choosing a sparse i.e, a supergaussian (heavy tailed) prior on $x$.
Choice of distributional parameter $p$ along with different suitable mixing densities, i.e $p(\gamma),$ will lead to different distributions including the super gaussian distributions.
Because of the scale mixture representation, the generation of the random variable $X$ can be viewed in a hierarchy,  i.e. generate $\gamma$ using $p(\gamma)$ followed by generating $X$ using $p(x|\gamma)$. The framework allows for dealing with complicated models in a simple manner and is indispensable as we move towards complex problems
with structure.

As special cases, the choice of $p=2$ leads to Gaussian Scale Mixtures (GSM) which has been very popular in the literature, and $p=1$ leads to the Laplacian Scale Mixtures (LSM).
Interestingly, a Laplacian distribution $p(x) = \frac{a}{2} e^{-a|x|}$ can be represented as a GSM with exponential mixing density $p(\gamma),$  i.e.
$p(\gamma) = \frac{a^2}{2} \exp(-\frac{a^2}{2} \gamma) u(\gamma),$  where $u(.)$ is the unit step function \cite{babacan2010bayesian}. More explicitly,
\begin{equation}
\begin{split}
p(x)&= \int_0^{\infty} p(x|\gamma)p(\gamma) d\gamma \\
&=\int_0^{\infty} \frac{1}{\sqrt{2 \pi \gamma}} \exp( - \frac{x^2}{2 \gamma}) \times \frac{a^2}{2} \exp(-\frac{a^2}{2} \gamma) d \gamma\\
& = \frac{a}{2} e^{-a|x|}
\end{split}
\label{laplacian}
\end{equation}
This means, any LSM can also be represented as a GSM with an extra layer of hierarchy. This will play an important role in the SSR algorithm development. This fact also leads to the observation of the relationship between the different scale mixture families as depicted in Figure~\ref{fig:venn}.

\subsection{An example of PESM: Generalized t Distribution}
\label{sec:t distribution}

In this example, we will consider an inverse generalized gamma (GG) distribution as our mixing density $p(\gamma)$ in the hierarchical representation (\ref{eq:mix}) for the PESM family. It leads to a generalized t distribution which is a superset of all the sparse distributions that have been used in practice in several recent works, e.g. Generalized double Pareto (GDP), Laplacian and Student-t distributions, among others.

The Generalized t Distribution has the form:

\begin{equation}
GT(x; \sigma, p, q) = \frac{\eta}{(1+\frac{|x|^p}{q\sigma^p})^{q+\frac{1}{p}}}
\end{equation}

Where $\eta$ is the normalization constant, $p$ and $q$ are the shape parameters and $\sigma$ is the scale parameter. Interestingly, $p$ and $q$ provide the flexibility to represent different tail behavior using this distribution. Larger values of $p$ and $q$ correspond to thin tailed distributions whereas smaller values of $p$ and $q$ are associated with heavy tailed distributions.

As mentioned above, the GT distribution family can  be represented in PESM framework using $p(\gamma) = GG(\gamma; -p, \sigma, q)$ where,

\begin{equation}
GG(x; -p, \sigma, q) = \eta \; (\sigma/x)^{pq+1} e ^{-(\sigma/x)^p}
\end{equation}

Interesting special case of note is  $p=2,$ which leads to a student t distribution, a prior that has been used in the popular Sparse Bayesian Learning (SBL)/Relevance Vector Machine (RVM) work and can be decomposed as a Gaussian Scale mixture with inverse Gamma as the mixing density. Employing $p=1$ leads to a Generalized Double Pareto distribution (GDP) discussed in \cite{armagan2013generalized} which can be represented as a scale mixture of Laplacian following equation \ref{eq:mix}.

In  Table \ref{tab:GT}, we summarize some special cases that have been used for SSR that arise by different choices of the shape parameters of GT, i.e. $p$ and $q.$

\vspace{1em}
%\textbf{Why PESM over GSM?}

%\vspace{1em}

Among Scale Mixtures, GSM in particular has gained a lot of interest over the years in the literature and  the proposed PESM framework is an interesting generalization for SSR purposes. 
As shown in \cite{palmer2005variational}, GSM can only be used to represent supergaussian densities, i.e. distributions with positive kurtosis whereas PESM representation can also be used for subgaussian densities along with supergaussian densities. One example is the previously discussed generalized t distribution, which becomes a thin tailed subgaussian distribution for $p>2$ ($q=1$, $\sigma=1$). Moreover, for the purposes of the SSR work, the general PESM allows one to treat both the
LSM and GSM in a unified manner thereby enabling treatment of $\ell_1$ and $\ell_2$ based algorithms in a unified manner.

\section{B-SSR: Type I}

Type I inference corresponds to standard MAP estimation technique in B-SSR. In this section we review the Type I framework and derive a Type I algorithm using PESM as the sparse prior. Then we specialize the result
using the Generalized t distribution as the sparse prior and also show that the generalized algorithm reduces to  well known SSR algorithms.

\subsection{Background on MAP Estimation (Type I methods)}
\label{sec:map}

Having chosen a sparsity enforcing distribution $p(\xbf),$
thereby allowing one to narrow the space of candidate solutions in
a manner consistent with application-specific assumptions, a
maximum a posteriori (MAP) estimator of $\xbf$ is then obtained as (Type I estimation)
\begin{equation}
\begin{split}
 \label{eq:map} \hat{\xbf} &= \arg \max_\xbf p(\xbf|\mathbf{y}) = \arg \max_\xbf p(\mathbf{y}|\xbf)p(\xbf) \\
  &=\arg \max_\xbf \left[ \log p(\mathbf{y}|\xbf) + \log p(\xbf) \right]
\end{split}
\end{equation}
Using the Gaussian noise assumption,
and a separable prior distribution $p(\mathbf{x})= \prod_i p(x_i)$, the MAP estimate is obtained by minimizing
\begin{equation}  J(\xbf) = \|\mathbf{\Phi} \xbf - \mathbf{y}\|^2_2 + \lambda
\sum_i g(x_i),
\label{eq:TypeI}
\end{equation}
where $g(x)$ is determined by $\log p(x).$
Incorporating sparsity by enforcing a sparse (supergaussian) distribution as the prior, $p(\mathbf{x}),$ reduces to  choosing $g(.)$. It has been shown that $g(.)$ which is symmetric, concave and nondecreasing functions on $[0, \infty)$ are useful choices in this context  \cite{palmer2010strong}.
Now, as discussed above, many of these sparse priors can be represented in a hierarchy and belong to the PESM family.

In order to contrast with the Type II formulation to follow,
with the PESM representation one can revisit the equation (\ref{eq:map}) and note that Type I involves integrating out the hyperparameter $\bm{\gamma}$.
\begin{equation}
\begin{split}
\hat{\mathbf{x}}&= \arg \max_x p(\mathbf{x}|\mathbf{y})\\ &= \arg \max_x p(\mathbf{y}|\mathbf{x}) \int p(\mathbf{x}|\bm{\gamma})p(\bm{\gamma}) d\bm{\gamma}
\end{split}
\end{equation}

\begin{table*}[t]
	\centering
	\caption{Variants of GT distribution and their connection to Type I Algorithms}
	\begin{tabular}{l c c c c} \\ [-2ex]
	\hline\hline \\
     	  q &p &Prior Distribution  & Penalty Function &SSR Algorithm  \\
     	   \hline \\
      $q \to \infty$  &2 &Normal  & $||x||_2$  &Ridge Regression\\
      $q \to \infty$  &1 &Laplacian   &$||x||_1$ &LASSO\\
     	$q\geq 0$ (degrees of freedom) &2 &Student t distribution  &$\log (\epsilon + x^2)$ &Reweighted $\ell_2$ (Chartrand's)\\
 $q\geq 0$ (shape parameter) &1 &Generalized Double Pareto  &$\log (\epsilon + |x|)$  &Reweighted $\ell_1$(Candes's)\\
     	   \hline \\ [-2ex]

	\end{tabular}
	\label{tab:GT}
\end{table*}

\subsection{Unified Type I Inference Procedure}
\label{type1_general}

In this section we derive the EM inference procedure for the PESM family in the Type I framework, i.e, we find the MAP estimate of $\mathbf{x}$ where a PESM has been employed for the sparsity inducing prior $p(\mathbf{x})$. Because of the separable prior, the $p(x_i)$  have an independent scale mixture representation,

\begin{equation}
p(x_i) = \int_0^{\infty} p(x_i|\gamma_i) p(\gamma_i) d\gamma_i
\end{equation}

For MAP estimation of $\mathbf{x},$ we treat the $\gamma_i$'s as hidden variables and employ an EM algorithm.  The complete data log-likelihood can be written as,

\begin{equation}
\log p(\mathbf{y}, \mathbf{x}, \bm{\gamma})= \log p(\mathbf{y}|\mathbf{x}) + \log p(\mathbf{x}|\bm{\gamma}) + \log p(\bm{\gamma})
\label{eq:likelihood1}
\end{equation}

To formulate the Q function, we need to find the conditional expectation of the complete data log-likelihood with respect to posterior of the hidden variables $p(\bm{\gamma}| \mathbf{x}, \mathbf{y})$ which reduces to $p(\bm{\gamma}| \mathbf{x})$ by virtue of the Markovian property induced by the hierarchy, i.e. ${\bm{\gamma}} \rightarrow \mathbf{x} \rightarrow \mathbf{y}.$
Since in the M step we need to maximize the Q function with respect to $\mathbf{x}$, we are only concerned with the first two terms in (\ref{eq:likelihood1}) and only the second term has dependencies on $\gamma_i$. This is the only term we need to be concerned with during the E-step.
Now from the scale mixture decomposition and considering the ith component of $\mathbf{x}$,
\begin{equation}
\log p(x_i|\gamma_i) = \log PE(x_i; p, \gamma_i) = -\frac{|x_i|^p}{\gamma_i^p} + \text{constants}
\end{equation}
Hence, for determining the Q function we need the following conditional expectation, $E_{\gamma_i|x_i} \big [ \frac{1}{\gamma_i^p}\big ]$.

To compute the concerned expectation we will use the following trick. Differentiating inside the integral of the marginal $p(x_i)$,
\begin{equation}
\begin{split}
p^\prime(x_i) &= \frac{d}{dx_i} \int_0^{\infty} p(x_i|\gamma_i) p(\gamma_i) d\gamma_i\\
&=- p \times |x_i|^{p-1} \text{sign} (x_i) \int_0^{\infty} \frac{1}{\gamma_i^p}  p(x_i,\gamma_i) d\gamma_i\\
&=- p \times |x_i|^{p-1} \text{sign} (x_i) p(x_i) \int_0^{\infty} \frac{1}{\gamma_i^p}  p(\gamma_i|x_i)d\gamma_i \\
&= -p \times |x_i|^{p-1} \text{sign} (x_i) p(x_i) E_{\gamma_i|x_i} \big [ \frac{1}{\gamma_i^p}\big ]
\end{split}
\end{equation}
Hence,
\begin{equation}
E_{\gamma_i|x_i} \big [ \frac{1}{\gamma_i^p}\big ] =  - \frac{p^\prime(x_i)}{p \times |x_i|^{p-1} \text{sign} (x_i) p(x_i)}
\end{equation}
and enables determining the Q function.
Then the M step reduces to,

\begin{equation}
\hat{\mathbf{x}}^{(k+1)} = \arg \min_{\mathbf{x}} \frac{1}{2 \sigma^2} || \mathbf{y} - \Phi \mathbf{x}||^2 + \sum_i  w_i^{(k)} |x_i|^p
\end{equation}
Where $\sigma^2$ is the variance of the measurement noise and $w_i^{(k)} = E_{\gamma_i|x_i^{(k)}}\big [ \frac{1}{\gamma_i^p}\big ] $.

Following the traditional path of EM, the algorithm is an iterative one, i.e, in the E step the weights are computed and in the M step a weighted norm minimization is solved. This alternate
procedure is carried out iteratively till convergence.

\subsection{Special cases of Type I using Generalized t distribution}
\label{type1_special}

In this section we  specialize the  derived unified Type I EM algorithm with the generalized t distribution as $p(x_i)$.
We can write $p(x_i) \sim exp(-f(x_i))$ where,
\begin{equation}
f(x_i) = (q+1/p) \log (1+ \frac{|x_i|^p}{q \sigma^p})
\end{equation}
Thus,
\begin{equation}
E_{\gamma_i|x_i} \big [ \frac{1}{\gamma_i^p}\big ] = \frac{f^{\prime}(x_i)}{p \times |x_i|^{p-1} \text{sign} (x_i)}
\end{equation}
Substituting the value of $f^{\prime}(x_i)$ we get,

\begin{equation}
E_{\gamma_i|x_i} \big [ \frac{1}{\gamma_i^p}\big ] = \frac{q+1/p}{ q \sigma^p + |x_i|^p}
\label{eq:TypeI_weights}
\end{equation}

So the M step will become,

\begin{equation}
\hat{\mathbf{x}}^{(k+1)} = \arg \min_{\mathbf{x}} \frac{1}{2 \sigma^2} || \mathbf{y} - \Phi \mathbf{x}||^2 + \sum_i  w_i^{(k)} |x_i|^p
\label{eq:gttype1}
\end{equation}
Where $\sigma^2$ is the variance of the measurement noise and $w_i^{(k)} = E_{\gamma_i|x_i^{(k)}} \big [ \frac{1}{\gamma_i^p}\big ]=\frac{q+1/p}{ q \sigma^p + |x_i^{(k)}|^p}$.

In following subsections we will show how with specific choices of the distribution parameters of the generalized t, we can derive well known Type I (MAP estimation) based SSR algorithms.
\vspace{1em}
\subsubsection{LASSO ($\ell_1$-minimization) \cite{tibshirani1996regression}}
\vspace{1em}

 Interestingly we see from Table I that for specific values of the shape parameters ($q \to \infty$ and $p=1$), a generalized t distribution can be used to represent a double exponential or Laplacian distribution. Now to relate with the unified Type I MAP estimation inference procedure, taking the limit as $q \to \infty$ and $\sigma=1$ in (\ref{eq:TypeI_weights}), we get $w_i =1$. Hence in the M step we are just solving a $\ell_1$ penalized regression once as the weights are not changing over iterations, which is essentially the LASSO algorithm.
\vspace{1em}
\subsubsection{Reweighted $\ell_1$-minimization (Candes et al \cite{candes2008enhancing})}

\vspace{1em}

The popular reweighted $\ell_1$-minimization (Candes et al \cite{candes2008enhancing}) is a special case of the MAP estimation approach using a generalized t distribution as sparse prior.

Selecting the parameters of the generalized t as follows; $ q=\epsilon, p=1, \sigma=1$, one obtains,
  \begin{equation}
p(x_i| \epsilon) = GT(1, 1, \epsilon)=  \frac{\eta}{\big(1+ \frac{|x_i|}{\epsilon}\big)^{(\epsilon+1)}}
\end{equation}
which when substituted in equation (\ref{eq:TypeI}),
results in the following cost function,
\begin{equation}
\label{eq:candes}
\min_x ||y- \Phi x||_2^2 + \lambda\sum_i \log(|x_i| + \epsilon)
\end{equation}

In \cite{candes2008enhancing}, the above mentioned cost function is optimized using a MM approach.
 Now substituting the distribution parameters in equation (\ref{eq:TypeI_weights}), the weights reduce to  $w_i = \frac{1+\epsilon}{\epsilon + |x_i|}$. These are the same weights obtained in \cite{candes2008enhancing} via a MM method and $p = 1$ in Equation (\ref{eq:gttype1}) results in a  weighted $\ell_1$ minimization problem with the weights being a function of the previous estimate. This special case of GT has been also called the Generalized Double Pareto (GDP) distribution in the literature \cite{armagan2013generalized}.

Following the scale mixture decomposition of the GT distribution, as shown in Equation (\ref{eq:mix}), since $p = 1$ we can represent the prior as a Laplacian Scale Mixture.

\begin{equation}
p(x)= \int p(x|\gamma) p(\gamma) d\gamma = \int \frac{1}{2 \gamma } e^{-\frac{|x|}{\gamma}} p(\gamma) d\gamma,
\end{equation}
where $p(\gamma) = GG(\gamma; -1, 1, \epsilon)$. This observation is summarized in the following lemma.
\vspace{0.5em}
\begin{lemma}
 Let $x\sim Laplacian(0, \gamma)$, $\gamma \sim  GG(\gamma; -1, 1, \epsilon))$, then the resulting marginal density for $x$ is $GT(1, 1, \epsilon)$.
 \end{lemma}
 \vspace{0.5em}

\subsubsection{Reweighted $\ell_2$-minimization (\cite{chartrand2008iteratively,rao1999affine})}
\vspace{1em}

Another popular SSR algorithm, the reweighted $\ell_2$ minimization can also be represented in a Bayesian Type I setting by employing a Student t distribution with degree of freedom $2\epsilon.$ This heavytailed sparse prior $p(x)$ is again a special case of the generalized t distribution as shown in the table.

\begin{equation}
p(x_i| \epsilon) = GT(\sqrt{2}, 2, \epsilon)= \frac{\eta}{\big(1+\frac{|x_i|^2}{2\epsilon}\big)^{(\epsilon+1/2)}}
\end{equation}
The nature of the tail of the student t distribution is controlled by degrees of freedom parameter $\epsilon$ and smaller values of $\epsilon$ correspond to heavier tails. The
associated diversity penalty factor is given by $g(x_i) = \log (x_i^2 + \epsilon)$.
 For a Type I inference procedure, we can utilize the unified approach discussed above in Section \ref{type1_special} and substitute the shape and scale parameters $p=2, q=\epsilon, \sigma= \sqrt{2}$ of the generalized t distribution in Equation (\ref{eq:TypeI_weights}) to obtain, $w_i = \frac{\epsilon +1/2}{2\epsilon + |x_i|^2}.$ Since $p = 2,$ Equation (\ref{eq:gttype1}) leads to the reweighted $\ell_2$ minimization algorithm as discussed in \cite{chartrand2008iteratively}.

\vspace{1em}
\section{B-SSR: Type II (Evidence Maximization)}
The success of Type II approaches like SBL for SSR problems  motivate the Type II approach for the general PESM family.
As special cases, the three Type I algorithms discussed in Section~\ref{type1_special} are explored in the Type II setting.
We also analyze the difference between a Type I algorithm and its Type II counterpart which provides insights into the reasons for superior recovery performance
of Type II methods.

In a Type II procedure, instead of integrating out the hypeparameters ${\bm{\gamma}}$, we estimate them using an evidence maximization method, i.e.
\begin{equation}
\label{type2}
\begin{split}
\hat{\bm{\gamma}} &= \arg \max_{\bm{\gamma}} p(\bm{\gamma}|\mathbf{y}) = \arg \max_{\bm{\gamma}} p(\bm{\gamma}) p(\mathbf{y}|\bm{\gamma})\\&= \arg \max_{\bm{\gamma}} p(\bm{\gamma})\int p(\mathbf{y}|\mathbf{x})p(\mathbf{x}|\bm{\gamma}) d \mathbf{x}
\end{split}
\end{equation}
The evidence framework integrates over the coefficient vector $\xbf$ to obtain the evidence $p(\ybf|\bm{\gamma})$. This evidence is weighted by the hyperprior $p(\bm{\gamma})$ and maximized over $\bm{\gamma}$.
Once $\bm{\gamma}$ is obtained, the relevant posterior $p(\mathbf{x}|\mathbf{y})$ is approximated, often as $p(\xbf|\ybf;\hat{\bm{\gamma}}),$ and the mean of the approximated posterior is used as a point estimate.
Sparsity is achieved by many of the $\gamma_i$ being zero \cite{tipping2001sparse,wipf2010iterative,wipf2008new}.

\subsection{Unified Type II EM algorithm}
\label{type2ssr}
To solve the above mentioned optimization problem, we again employ the EM algorithm this time by treating $\mathbf{x}$ as the hidden variable. As in Section~\ref{type1_general}, we assume
a sparse prior $p(\mathbf{x})$ from the PESM family has been utilized and that the measurement noise is Gaussian with variance $\sigma^2$.

Hence the Q function has the form,
\begin{equation}
\begin{split}
Q(\bm{\gamma})&=E_{\mathbf{x}|\mathbf{y}; \bm{\gamma},  \sigma^2} [ \log p(\mathbf{y}|\mathbf{x}) + \log p(\mathbf{x}|\bm{\gamma}) + \log p(\bm{\gamma})]\\
&\approx E_{\mathbf{x}|\mathbf{y}; \bm{\gamma},  \sigma^2} [\sum_i -\frac{1}{p}\log {\gamma_i} - \frac{|x_i|^p}{\gamma_i} + \log p(\gamma_i) ]
\end{split}
\label{qfunction}
\end{equation}
Since in the M step we are only concerned with the terms involving $\bm{\gamma}$, examining them reveals that the E-step requires the computation of the following conditional expectation
\begin{equation}
\label{eq:estep}
E_{\mathbf{x}|\mathbf{y}; \bm{\gamma}^t,  {\sigma^2}} [|x_i|^p] = <|x_i|^p>
\end{equation}
In the M step we will maximize the Q function with respect to $\gamma_i$ to find the update rules.
To illustrate, if we consider a non informative hyperprior, i.e, $p(\gamma_i)=1$,
\begin{equation}
Q(\bm{\gamma})=\sum_i -\frac{1}{p}\log {\gamma_i} - \frac{<|x_i|^p>}{\gamma_i}
\end{equation}
Taking the derivative of the Q function w.r.t $\gamma_i$ and setting it to zero results in,
\begin{equation}
\hat{\gamma_i} = p <|x_i|^p>
\end{equation}
Since the E step requires the computation of the conditional expectation given by Equation (\ref{eq:estep}), we can either look for a closed form solution or revert to the MCMC technique \cite{park2008bayesian}.
We will examine this further for some special cases later.

\vspace{0.5em}
\subsection{Difference between Type I and Type II inference methods}

Type I and Type II provide two different approaches to solving the SSR problem. Hence it is important to understand  the theoretical differences between the  two inference procedures
to identify their suitability for SSR. In \cite{wipf2004perspectives}, the authors provide evidence for SBL, using a variational approximation to the  prior $p(\xbf),$
that Type II methods attempt to approximate the true posterior $p(\xbf|\ybf).$
If the true posterior distribution has a skewed peak, then the type I estimate (MAP of $\mathbf{x}$) is not a good representative of the whole posterior.  By trying to approximate the true posterior mass,
Type II methods are likely to provide a better estimate. Similar discussion of Type II desirability is provided in \cite{mackay1999comparison} in the context of general Bayesian inferencing.
We revisit the issue and attempt to corroborate this by exploiting specific attributes of the SSR problem. We first manipulate the Type II objective as shown below.
\begin{equation}
\begin{split}
p(\bm{\gamma}|\ybf)&= \int p(\bm{\gamma}, \xbf| \ybf) d\xbf\\ &= \int p(\bm{\gamma} | \xbf, \ybf) p(\xbf|\ybf) d\xbf\\ &=\int p(\bm{\gamma} | \xbf) p(\xbf|\ybf) d\xbf\\
 &= p(\bm{\gamma}) \int \frac{p(\xbf | \bm{\gamma})}{p(\xbf)} p(\xbf | \ybf) d\xbf
 \end{split}
\end{equation}
Lets assume that  $\bm{\hat{\gamma}}$ is the solution of Equation (\ref{type2}). It will be sparse for specific choice of $p(\bm{\gamma})$ as shown in \cite{wipf2010iterative, wipf2008new}.

Now, let $\underline{S}$ be the index of non zero entries and $\overline{S}$ be the index of zero entries. So, we can say $\bm{\hat{\gamma}}_{\overline{S}}=0$.
\begin{equation}
\begin{split}
p(\hat{\bm{\gamma}}| \ybf) &= \lim_{\epsilon \to 0} p(\hat{\bm{\gamma}} + \epsilon | \ybf)\\
&= p(\hat{\bm{\gamma}}) \lim_{\epsilon \to 0} \int_{\underline{S}} \int_{\overline{S}} \frac{p(\xbf_{\underline{S}} | \hat{\bm{\gamma}_{\underline{S}}} + \epsilon_{\underline{S}})p(\xbf_{\overline{S}} | \epsilon_{\overline{S}})}{p(\xbf_{\underline{S}})p(\xbf_{\overline{S}})} p(\xbf|\ybf) d\xbf
\end{split}
\end{equation}
$ p(\xbf_{\overline{S}}| \epsilon_{\overline{S}})$ is a normal distribution with mean zero and variance $\epsilon_{\overline{S}}$. Hence when $\epsilon_{\overline{S}} \to 0$, $ p(\xbf_{\overline{S}}| \epsilon_{\overline{S}})$ becomes a dirac delta function, i.e. $\delta(x_{\overline{S}})$.

Using the properties of dirac delta functions inside the integration, we obtain
\begin{equation}
\begin{split}
p(\hat{\bm{\gamma}} | \ybf) &= \int_{\underline{S}} \frac{p(\xbf_{\underline{S}}|\hat{\bm{\gamma}}_{\underline{S}})}{p(\xbf_{\underline{S}})}  \frac{p(\hat{\bm{\gamma}})}{p(\xbf_{\overline{S}}=0)} p(\xbf_{\underline{S}}, \xbf_{\overline{S}}=0|\ybf) d\xbf_{\underline{S}}
\end{split}
\end{equation}
Hence from this analysis, we see that we are evaluating a weighted integral of the true posterior $p(\xbf|\ybf)$ over the subspaces spanned by the non zero indexes.
This shows that in the evidence maximization framework instead of looking for the mode of the true posterior $p(\xbf|\ybf)$, we approximate the true posterior by $p(\xbf|\ybf; \hat{\bm{\gamma}})$ where $\hat{\bm{\gamma}}$ is obtained by maximizing the true posterior mass over the subspaces spanned by the non zero indexes.
This is in contrast to Type I methods that seek the mode of the true posterior and use that as the point estimate of the desired coefficients.
\vspace{0.5em}

%\textbf{Robustness of Type II (Hierarchical) framework:}

%\vspace{0.5em}
Another favorable aspect of the
Type II framework is that it inherits the robustness property of a Hierarchical Bayesian modeling framework. It has been shown extensively in the statistics literature \cite{lehmann1998theory, gustafson1996aspects, goel1981information}, that the posterior of  a hyperparameter, i.e, $\bm{\gamma},$ is less affected by the wrong choices of prior than the posterior of the parameter  $\mathbf{x}$. In other words, parameters that are deeper in the hierarchy have less effect on the inference procedure, which allows us to be less concerned about the choice of $p(\bm{\gamma})$. Another virtue is that the hierarchical framework allows for parameter tying and this can greatly
reduce the search space for Type II methods by leading to an optimization problem with fewer parameters. This is more evident for problems like the MMV and block sparsity problem \cite{zhang2011sparse, wipf2007empirical,zhang2010sparse}.


\subsection{Special case of Unified Type II with different choices of $p$}
\label{special_TypeII}

As discussed above for the unified Type II approach our concerned posterior is $p(\mathbf{x}|\mathbf{y}; \bm{\gamma}, \sigma^2)$.
For a point estimate of $\mathbf{x}$ we will use the mean of the posterior, $\hat{\mathbf{x}} = \int \mathbf{x} p(\mathbf{x}|\mathbf{y}; \bm{\gamma}, \sigma^2)d \mathbf{x}$.
Now the posterior could be computed as,
\begin{equation}
\begin{split}
p(\mathbf{x}|\mathbf{y}; \bm{\gamma},  {\sigma^2}) &\approx p(\mathbf{y}|\mathbf{x}) p(\mathbf{x}|\bm{\gamma})\\
&\approx \exp \huge\{ -\frac{1}{2\sigma^2} ||\mathbf{y} - \Phi \mathbf{x}||_2^2 - \sum_i \frac{|x_i|^p}{\gamma_i}\huge\}
\label{post}
\end{split}
\end{equation}
The challenge is proper normalization and tractability of the computation of the mean. For the EM algorithm to be successfully implemented, one must also
be able to carry out the E-step, Equation (\ref{eq:estep}). We now explore this for some specific PESM family members.

\subsubsection{Choice of $p=2$}

Choice of $p=2$ corresponds to Gaussian Scale Mixture, and is very tractable. The GSM based Type II methods have been extensively studied \cite{ji2008bayesian, tipping2001sparse, wipf2004perspectives} and so we keep the discussion brief.
This choice (in Equation (\ref{post})) leads to a Gaussian posterior  given by
\begin{equation}
p(\mathbf{x}|\mathbf{y}; \bm{\gamma}, \sigma^2)= N(\mu , \Sigma)
\label{estimate}
\end{equation}
where \begin{equation}
\label{eq:mean}
 \mu=  \Gamma \Phi^T (\sigma^2 I + \Phi \Gamma \Phi^T)^{-1}\ybf
 \end{equation}
 \begin{equation}
\Sigma= \Gamma- \Gamma \Phi^T (\sigma^2 I + \Phi \Gamma \Phi^T)^{-1}\Phi \Gamma \label{eq:sigma}
\end{equation}
and $\Gamma = \text{diag}(\bm{\gamma})$.
The EM algorithm can also be readily carried out because the E-step requires the second moment which can be readily obtained using Equation (\ref{eq:sigma}).
The estimate of $\bm{\gamma}$ in the M step and the updates of $\bm{\gamma}$ depend on the mixing density $p(\bm{\gamma})$ as shown in Equation (\ref{qfunction}) and can be readily carried out for the non-informative prior
and for a reasonable large class of priors \cite{palmer2005variational}.
The true posterior can be approximated by a Gaussian distribution whose mean and covariance depend on the estimated hyperparameters. Now, for a point estimate of the coefficient vector, we will choose,
\begin{equation}
\mathbf{\hat{x}} = \mu.
\end{equation}
From Equation~(\ref{eq:mean}), one can see that $\mu$ is sparse if $\bm{\gamma}$ is sparse.
To complete the discussion, we discuss the most popular of the Type II methods.
In Relevance Vector Machine (Type II) \cite{tipping2001sparse}, Tipping has shown that the 'true' coefficient prior used in SBL actually follows a student t distribution (GSM with Gamma distribution as mixing density), and discusses in detail how the hierarchical formulation of this prior helps to realize the supergaussian nature. Hence we can see that the corresponding Type II formulation of Reweighted $\ell_2$ is SBL with a slight difference. In SBL $\epsilon$ is set to zero which gives us an improper prior $ p(x) \sim 1/{|x|}$ which is sharply peaked at zero. But as discussed in previous literatures, $\epsilon =0$ in Type I version, i.e, in Reweighted $\ell_2$  increases the number of local minima and convergence to a sub optimal solution becomes more likely.
Now to solve the M step for this case we will use the following PESM ($p=2$) formulation,
\vspace{0.5em}

\begin{lemma}
 Let $x\sim N(0, \gamma)$, $\gamma \sim Inverse-Gamma( \epsilon, \epsilon)$  Then the resulting marginal density for $x$ is $GT(\sqrt{2}, 2, \epsilon)\simeq Student-t(2\epsilon)$.
\end{lemma}

Details of this inference procedure can also be found in \cite{tipping2001sparse, ji2008bayesian}, and update rules have been shown in Table II.

\subsubsection{Choice of $p=1$}

With $p=1,$ PESM reduces to a Laplacian Scale Mixture. To successfully carry out the EM algorithm, the E-step requires
the computation of $E(|x_i| ;\ybf,\gamma^{(k)}).$ A closed form expression does not appear feasible and a more numerical approach may be required.
Also, the concerned posterior (Equation \ref{post}) does not appear to have a simple closed form expression making final inferencing a challenge
along with the computation of the mean for the point estimate. An efficient numerical approach needs to be developed and is left for future work.

In this work, we follow an alternate strategy and take advantage of the fact that the LSM family is contained within the GSM family.
Since a Laplacian distribution can be written as a member of the GSM family (Equation \ref{laplacian}) \cite{babacan2010bayesian,figueiredo2003adaptive}, it will be possible to get a closed form posterior using a three layer hierarchy. We will illustrate this for the prior associated with  Type I Reweighted $\ell_1$-minimization approach and develop a Type II variant.
The closed form posterior will be Gaussian and have the same form for the case of $p=2$ as shown in Equation (\ref{estimate}). The only difference between $p=2$ and $p=1$ lies in the estimation of the hyperparameters.
%$\gamma$, i.e in the M step because of different choices of $p(\gamma)$.

Type II $\ell_1$  variant can also be derived and has been dealt with in previous work \cite{babacan2010bayesian} and for sake of completeness the update rule is summarized in Table II
along with other Type II algorithms. We will now derive the M step for the case of Type II Reweighted $\ell_1$-minimization which can be
followed in a straightforward manner for other cases including the $\ell_1$ variant.

We have shown in the discussion of Type I Reweighted $\ell_1$ that the concerned prior $GT(1,1,\epsilon)$ in a Bayesian setting is a Laplacian Scale mixture. This prior  can  be represented in a 3 layer hierarchy involving a GSM representation for the Laplacian density as summarized below.
\vspace{0.5em}

\begin{lemma}
 Let $x\sim N(0, \gamma)$, $\gamma \sim Exp(\frac{\lambda^2}{2})$ and  $\lambda \sim Ga(\epsilon, \epsilon)$ where $\epsilon>0$. Then the resulting marginal density for $x$ is $GT(1, 1, \epsilon)$.
\end{lemma}

\vspace{0.5em}
\begin{figure}
\centering
  \includegraphics[width=\linewidth,height=2.0in]{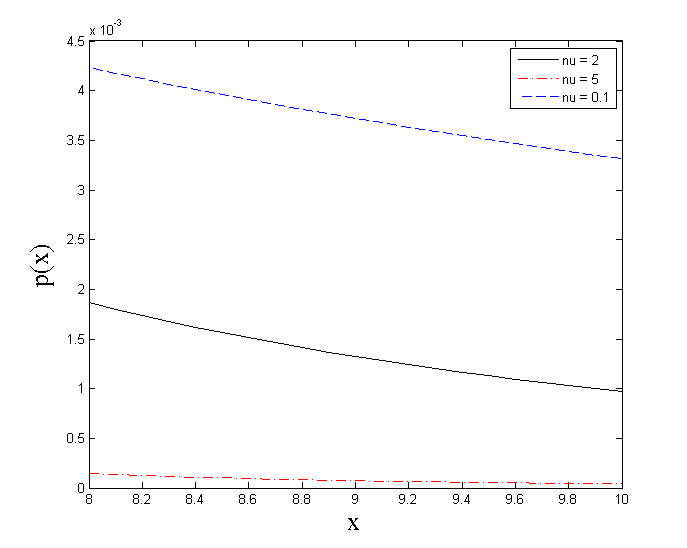}
  \caption{Tail behavior of Student's t distribution for different values of degrees of freedom}
  \label{pdf}
\label{fig:test1}
\end{figure}

\vspace{0.5em}

\begin{figure}
\centering
  \includegraphics[width=\linewidth,height=2.0in]{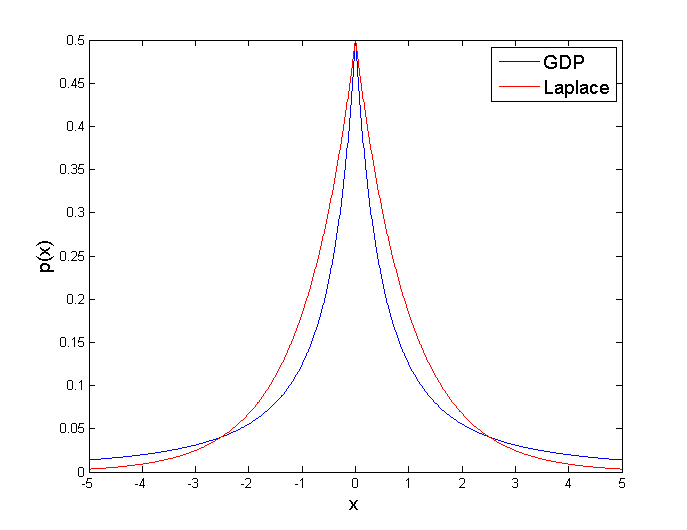}
  \caption{Comparison of tail behavior of two distributions: Generalized Double Pareto (GDP) and Laplacian}
  \label{pdf}
\label{fig:test}
\end{figure}

Fig. \ref{fig:test} compares two corresponding densities, $GT(1,1,1)$ and Laplace distribution with $\lambda=1$. It is evident from this figure that the Laplace prior has relatively light tails which contributes to the problem of over-shrinking of the large coefficients. On the other hand, the generalized t distribution has relatively heavier tails and a peak at zero which promotes zero coefficients. This is another reason of the superior recovery performance of Reweighted $\ell_1$-minimization over the LASSO, i.e. $\ell_1$-minimization, approach.

\begin{table*}[t!]
	\centering
	\caption{Updating Hyperparameters of Type II Algorithms}
	\begin{tabular}{l c c c } \\
	\hline\hline \\
     	  Type II algorithm &Mixing Density  & Update Rules  \\
     	   \hline \\
        \small Type II $\ell_1$ &\small $p(\gamma_i|\lambda) = Exp(\lambda/2)$  & $\hat{\gamma_i} = \frac{-1+\sqrt{1+4\lambda (\mu_i^2  + \Sigma_{i,i})}}{2\lambda}$, $\hat{\lambda} = \frac{2M}{\sum_i \gamma_i}$\\

     \small Type II Re-$\ell_1$ (Candes) &\small $p(\gamma_i|\lambda) = Exp(\lambda^2/2)$, $p(\lambda) = Gamma(\epsilon, \epsilon)$ & $\hat{\gamma_i} = \frac{-1+\sqrt{1+4\lambda^2 (\mu_i^2  + \Sigma_{i,i})}}{2\lambda^2}$, $\hat{\lambda} = \frac{-\epsilon+\sqrt{\epsilon^2+4(2M+\epsilon -1)\sum \gamma_i}}{2\sum \gamma_i}$\\
    \small Type II Re-$\ell_2$ (Chartrand) &\small $p(\gamma_i|\epsilon) = Inv-Gamma(\epsilon, \epsilon) $  & $\hat{\gamma_i} =\frac{\mu_i^2  + \Sigma_{i,i}+2\epsilon}{2\epsilon+1}$\\
     	   \hline \\ [-2ex]

	\end{tabular}
	\label{tab:reverb2014_compare}
\end{table*}

Now, for estimation of hyperparameters $\bm{\gamma}$ and $\lambda$ in the three layer hierarchy,
an EM algorithm will be developed. As in Section~\ref{type2ssr}, using $(\ybf,\xbf)$ as the complete data,
 maximizing the conditional expectation of the  complete data log likelihood involves maximizing,
\begin{equation}
\begin{split}
Q(\bm{\gamma}, \lambda, \sigma^2)&=E_{\mathbf{x}|\mathbf{y}; \bm{\gamma}, \lambda, \sigma^2} [ \log p(\mathbf{y}, \mathbf{x}; \bm{\gamma}, \lambda, \sigma^2)]
\end{split}
\end{equation}

In the E step, for iteration $t,$ we only need to compute the second moment which is straightforward because of the GSM representation of the Laplacian, i.e.
\begin{equation}
E_{\mathbf{x}|\mathbf{y}; \mathbf{\gamma}^t, \lambda, {\sigma^2}} [x_i^2] = \Sigma_{(i,i)}+ \mu_i^2
\end{equation}
\begin{figure}
\centering
  \includegraphics[width=\linewidth,height=2.0in]{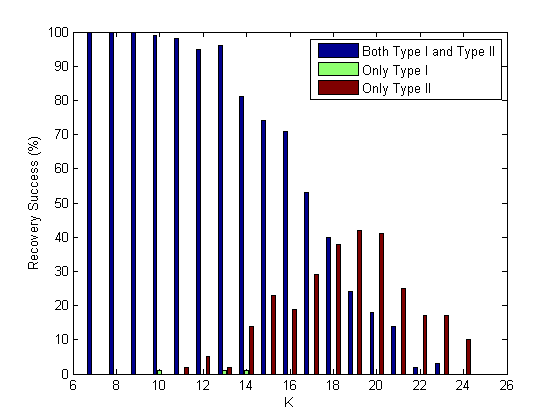}
  \caption{Bar plot of the recovery performance for Type I and
Type II Reweighted $\ell_1$ (Candes et al) minimization}
  \label{pdf}
\label{fig:test11}
\end{figure}
In the M step, the Q function is maximized with respect to the hyperparameters, $\bm{\gamma}$ and $\lambda$.

\begin{equation}
\begin{split}
Q(\bm{\gamma}, \lambda)&=E_{\mathbf{x}|\mathbf{y}; \bm{\gamma}, \lambda, \sigma^2} [ \log p(\mathbf{y}|\mathbf{x}) + \log p(\mathbf{x}|\bm{\gamma})\\
 &+ \log p(\bm{\gamma}|\lambda)+\log p(\lambda|\epsilon)]
\end{split}
\end{equation}
Now using the E step and only retaining the terms that involve $\bm{\gamma}$ and $\lambda$ we obtain,
\begin{equation}
\begin{split}
Q(\bm{\gamma}, \lambda)&= -\frac{1}{2} \sum_i \log \gamma_i - \frac{1}{2} \sum_i \frac{\Sigma_{(i,i)}+ \mu_i^2}{\gamma_i} \\
&+ \sum_i (2\log \lambda - \frac{\lambda^2}{2}\gamma_i) +(\epsilon-1) \log \lambda - \epsilon \lambda
\end{split}
\end{equation}

In the M step, taking the derivative of the Q function w.r.t $\gamma_i$ and $\lambda$ and setting to zero results in.

\begin{equation}
\frac{\partial Q}{\partial \gamma_i} = -\frac{1}{2\gamma_i} + \frac{\Sigma_{(i,i)}+ \mu_i^2}{2 \gamma_i^2} -\frac{\lambda^2}{2} = 0
\end{equation}
Solving this quadratic equation we obtain,
\begin{equation}
\hat{\gamma_i} = \frac{-1+\sqrt{1+4\lambda^2 (\mu_i^2  + \Sigma_{i,i})}}{2\lambda^2}
\end{equation}

Similarly,

\begin{equation}
\frac{\partial Q}{\partial \lambda} = \frac{2M+\epsilon-1}{\lambda} -\lambda \sum_i \gamma_i -\epsilon= 0
\end{equation}
Hence,
\begin{equation}
\hat{\lambda} = \frac{-\epsilon+\sqrt{\epsilon^2+4(2M+\epsilon-1) \sum_i \gamma_i}}{2\sum_i \gamma_i}
\end{equation}
We can also estimate the measurement noise variance $\sigma^2$ by maximizing the above Q function as shown in \cite{tipping2001sparse}. In this work, for simplicity, we will assume that the SNR of the environment is known to us before hand. We can also employ a fixed point optimization technique as shown in \cite{tipping2001sparse} to estimate the hyperparameters.

After convergence, one finds that most of the $\gamma_i,$  i.e. the variance of the normal distribution are driven to zero, which makes the associated coefficient zero and prunes it out from the model.

\begin{figure}
\centering
  \includegraphics[width=\linewidth,height=2.8in]{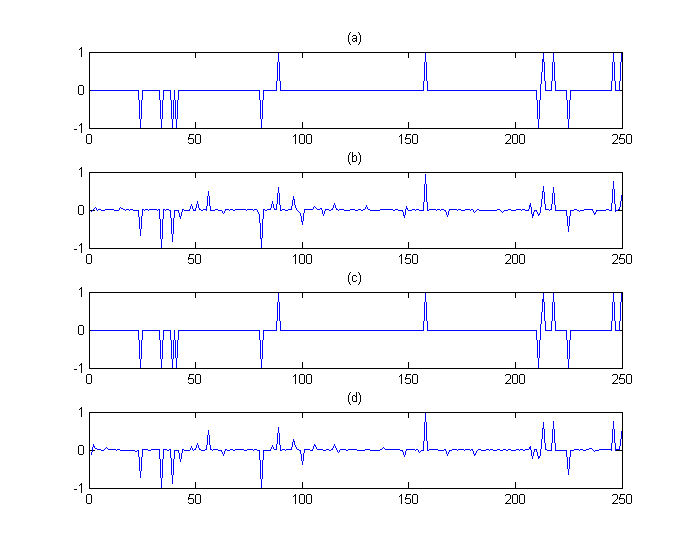}
  \caption{Reconstruction of uniform spikes where $k=13$ using (a) Original Signal, (b) $\ell_1$ norm minimization (Type I), (c) Type II $\ell_1$ minimization, (d) Candes et al (Type I) Reweighted $\ell_1$ minimization}
  \label{pdf}
\label{fig:test10}
\end{figure}
\section{Numerical Experiments}
\label{results}

In this section we present a set of experiments to evaluate and compare the Type II/Hierarchical framework based methods with those based on  regularization framework, i.e. Type I methods (MAP estimation), for the task of sparse signal recovery.
The experimental setup used is quite standard and has been used widely in  the SSR literatures.

\subsection{Problem Specification}
The measurement vector $\ybf$ is generated using a $N \times M = 50 \times 250$ dictionary $\Phi$, whose elements are generated from a i.i.d normal distribution with mean=0 and variance=1. A sparse signal $\mathbf{x}_{gen}$ of length 250
is generated such that $||\mathbf{x}_{gen}||_0=k$. The support, i.e. the location of the k nonzero elements, is chosen randomly, and the values are chosen from three different distributions:

\begin{itemize}
\item[(I)]
Uniform $\pm 1$ random spikes. (Sub-Gaussian)
\item[(II)]
Zero mean unit variance Gaussian.
\item[(III)]
Student t distribution with degrees of freedom $\nu=3$. (Super-Gaussian)
\end{itemize}

The synthetic measurements are generated using $\ybf = \Phi \mathbf{x}_{gen}$.  The generated measurements and the dictionary are then provided as input to the algorithms. The estimated coefficients are compared with the original $\mathbf{x}_{gen}$ that has been used to generate the measurement. For a single instance, the method is credited with a successful recovery if the estimate $\hat{\xbf}$ satisfies,
\begin{equation}
 ||\mathbf{x}_{gen}-\hat{\xbf}||_{\infty}\leq 10^{-3}
 \end{equation}
500 trials are conducted for various fixed combinations of k, i.e. the number of non zero coefficients, and the probability of successful recovery is plotted with respect to k. As expected, the probability of successful recovery decreases as  k, i.e. the cardinality of support, increases.

\subsection{Recovery Performance}

\subsubsection{Competing Algorithms}

Since the main goal of our work is to compare the Type I algorithms with their Type I counterparts, we designed the Type II versions of three well known norm minimization based Type I algorithms and compare their performance.
The algorithms in the study are:
\vspace{1em}
\begin{itemize}
\item $\ell_1$ minimization based SSR. (Basis Pursuit)
\item
Type II $\ell_1$ minimization based SSR. (Fixed $\lambda = 5$)
\item
Type I Reweighted $\ell_1$ minimization. ($\epsilon=0.1$ \cite{candes2008enhancing})

\item
Type II Reweighted $\ell_1$ minimization (Fixed $\epsilon =100$)

\item
Type I Reweighted $\ell_2$ minimization. ($\epsilon$ regularized, optimal update from \cite{chartrand2008iteratively})

\item
Type II Reweighted $\ell_2$ minimization (Fixed $\epsilon =0$: SBL)
\end{itemize}

\begin{figure}
\centering
  \includegraphics[width=\linewidth,height=2.2in]{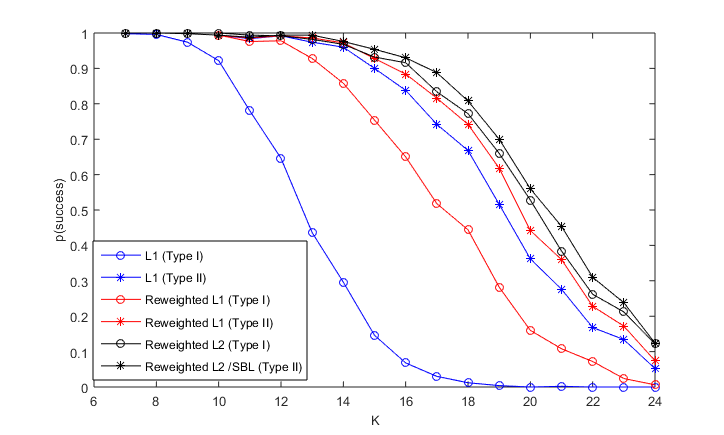}
  \caption{Recovery performance with Gaussian distributed non zero coefficients}
  \label{pdf}
\label{fig:test5}
\end{figure}

\begin{figure}
\centering
  \includegraphics[width=\linewidth,height=2.2in]{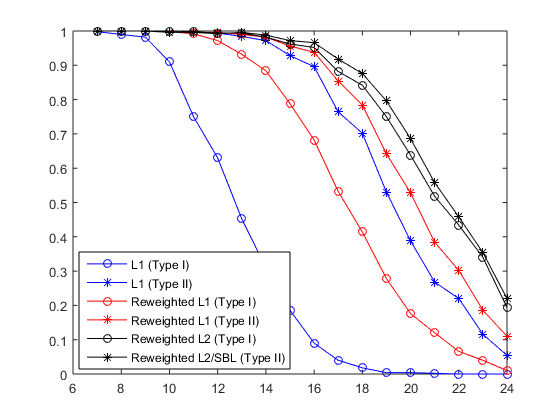}
  \caption{Recovery performance with Super Gaussian (Student t) distributed non zero coefficients}
  \label{pdf}
\label{fig:test6}
\end{figure}

\begin{figure}
\centering
  \includegraphics[width=\linewidth,height=2.2in]{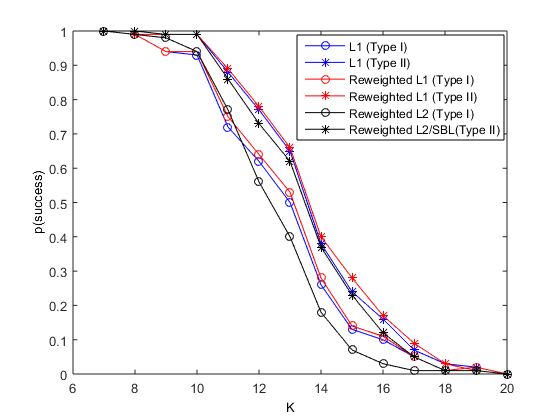}
  \caption{Recovery performance with Sub Gaussian distributed non zero coefficients}
\label{fig:test7}
\end{figure}

\vspace{0.5em}
\subsubsection{Performance Comparison}

In Figure \ref{fig:test5}, the probability of successful recovery with increasing support cardinality is plotted for the case where the non zero coefficients are from  a zero mean, unit variance, Gaussian distribution. It is evident from this plot that for all the algorithms, Type II versions outperform their Type I counterparts. This performance difference is significant in case of $\ell_1$ norm minimization.  Type I Reweighted $\ell_2$ minimization approach works much better compared to other two Type I methods, and the reason being the heuristic update of $\epsilon$, which helps it to get away from local minima. Hence, $\epsilon$ update in Reweighted $\ell_2$ (Type I) is absolutely necessary as we have found out for fixed $\epsilon$ this algorithm's performance decreases significantly.  Figure \ref{fig:test11} shows this comparison for the Reweighted $\ell_1$ minimization (Candes et al) in detail. The figure indicates trials when both Type I and Type II method have been successful and when only one of them has been successful and it is evident that for high values of $k,$ Type II outperforms Type I by a significant margin.

In Figure \ref{fig:test6}, the probability of successful recovery with increasing support cardinality is plotted where the non zero coefficients are generated from a student's t distribution with degrees of freedom 3. Again, the empirical superiority of the Type II versions over their Type I counterparts is evident from Figure \ref{fig:test6}. Interesting point to note here, is the performance improvement of Type I and Type II version of Reweighted $\ell_2$ algorithm over the others is significant and the reason could be that assumed prior for the non zero coefficients and the true prior have the same tail behavior (student's t) and are better matched.

Finally, we repeat the same set of experiments where the non zero coefficients follow a sub-gaussian distribution, i.e. Uniform $\pm 1$ random spikes, and the plot of the probability of successful recovery with increasing support cardinality is shown in Figure \ref{fig:test7}.  Though Type II methods still outperform their Type I counterparts, the performance improvement is less significant compared to the previous two cases. The reason could be that, since the assumed priors are supergaussian, i.e. heavy tails, it is difficult to model the true prior, i.e. sub gaussian density, for the nonzero coefficients. In Figure \ref{fig:test10}, an instance of reconstruction is shown using $k=13$ along with the original signal. It is evident that both $\ell_1$ minimization (Type I) and Candes's Reweighted $\ell_1$ minimization (Type I) fail, whereas Type II version of $\ell_1$ minimization recovers the original signal. For this instance, the other three SSR algorithms have also been successful in recovering the original signal.

\section{Conclusion and Discussion}

In this paper, we formulated the SSR problem from a Bayesian perspective and presented two different Bayesian frameworks which encompass all the well known recovery algorithms in practice. We presented a generalized scale mixture family : PESM, which is of prime importance for the design of Hierarchical Bayesian Recovery algorithms, i.e, Type II algorithms. The unified treatment of both $\ell_1$ and $\ell_2$ norm minimization based algorithms along with the design of Type II version of the Reweighted $\ell_1$ minimization algorithm are the main contributions of this work.

We also showed that, in a hierarchical Bayes framework instead of looking for a mode of the true posterior Type II methods actually try to find an approximate posterior such that the mass of the true posterior over the subspace spanned by non zero indexes is maximized. This leads to a better approximation of the true posterior, which is the reason behind the superior empirical results obtained using the Type II framework. Type II framework also enjoys the robustness property inherited because of its connection with Hierarchical Bayes which allows one to be less concerned about the choice of prior on the hyperparameters.

\bibliographystyle{IEEEbib}
\bibliography{tsp_2015}
\end{document}